\documentclass{article} 

\usepackage[utf8]{inputenc} 

\usepackage{geometry} 
\usepackage{graphicx} 

\usepackage{booktabs} 
\usepackage{array} 
\usepackage{paralist} 
\usepackage{verbatim} 

\usepackage{fancyhdr} 
\usepackage{sectsty}
\allsectionsfont{\sffamily\mdseries\upshape} 

\usepackage[nottoc,notlof,notlot]{tocbibind} 
\usepackage[titles,subfigure]{tocloft} 

\usepackage{multirow,array}
\usepackage{float} 
\usepackage{amsmath}
\usepackage{amsfonts}
\usepackage{amsthm}

\let\origref \ref
\def \ref#1{\textbf{\origref{#1}}}

\usepackage{subcaption} 

\usepackage[natbibapa]{apacite}
\setcitestyle{authoryear,open={(},close={)}}

 \usepackage{tikz} 
\usepackage{tikzscale}
\usetikzlibrary{positioning,arrows}
   \tikzset{
   modal/.style={>=stealth,shorten >=1pt,shorten <=1pt,auto,node distance=1.5cm,
   semithick},
   world/.style={circle,draw,minimum size=0.5cm,fill=gray!15},
   point/.style={circle,draw,inner sep=0.5mm,fill=black},
   reflexive above/.style={->,loop,looseness=7,in=110,out=70},
   reflexive below/.style={->,loop,looseness=7,in=240,out=300},
   reflexive left/.style={->,loop,looseness=7,in=150,out=210},
   reflexive right/.style={->,loop,looseness=7,in=30,out=330}}
\usetikzlibrary{calc, shapes}

\usetikzlibrary{fit} 

\usetikzlibrary{shadows,arrows,positioning}
 \usepackage{tikz-network}

\usepackage{mathtools}

\DeclarePairedDelimiter\abs{\lvert}{\rvert}%
\DeclarePairedDelimiter\norm{\lVert}{\rVert}%

\makeatletter
\let\oldabs\abs
\def\abs{\@ifstar{\oldabs}{\oldabs*}}
\let\oldnorm\norm
\def\norm{\@ifstar{\oldnorm}{\oldnorm*}}
\makeatother

\usepackage{graphicx,multirow}

\usepackage[title,toc,titletoc]{appendix}

\usepackage{hyperref}

\title{Compositional Understanding in Signaling Games}
\author{
David Peter Wallis Freeborn 
\thanks{Simulation code available at: \href{https://github.com/DavidFreeborn/compositional-signals}{github.com/DavidFreeborn/compositional-signals}} \\
Department of Philosophy \\
Northeastern University London \\
\texttt{david.freeborn@nulondon.ac.uk} \\
\texttt{ORCID: 0000-0002-2117-8145}
}
\date{}

\begin{document}
\maketitle

\abstract{
 Receivers in standard signaling game models struggle with learning compositional information. Even when the signalers send compositional messages, the receivers do not interpret them compositionally. When information from one message component is lost or forgotten, the information from other components is also erased. In this paper I construct signaling game models in which genuine compositional understanding evolves. I present two new models: a minimalist receiver who only learns from the atomic messages of a signal, and a generalist receiver who learns from all of the available information. These models are in many ways simpler than previous alternatives, and allow the receivers to learn from the atomic components of messages.}

\section{Introduction}
\label{sec:intro}

Two-player signaling games provide a model in which arbitrary signals can acquire conventional meanings, shared between the different players \citep{skyrms2010signals, lewis1969convention}. However, sophisticated communication systems, such as human language, do not merely use signals in isolation. Instead, human language is widely held to be \emph{compositional}: we can generate complex expressions, whose meaning is determined by their constituent parts, and the rules used to combine them. 

Can signaling games help to explain the emergence of compositional communication? A number of recent studies \citep{barrett2006numerical, barrett2007conventionality, barrett2009coding,franke2016compositionality, Franke2014, scott2012communication, steinert2016compositional, steinert2020emergence} attempt to construct signaling game models in which, at first glance,  the receiver \emph{seems} to learn the meaning of compositional signals. 

However, some authors have questioned whether the communication systems learned in conventional signaling games models are truly compositional \citep{franke2016compositionality, Franke2014, steinert2016compositional, lacroix2023noncompositional}. As \citet[p.~362]{franke2016compositionality} puts it, ``there is no reason to assume that the component parts of complex signals are independently meaningful to the agents''. \citet{lacroix2023noncompositional} makes these arguments explicit, using information-theoretic reasoning. If certain messages are replaced or forgotten, information is lost from other messages in the signal. The loss of information suggests that the receiver could never have been interpreting the signal in a compositional way to begin with. LaCroix argues that such syntactic signaling cannot suffice to explain the evolution of compositionality \citep{lacroixlogictologic}. Whether or not this is right, these arguments point to a serious problem with conventional signaling games models. Ostensibly compositional models do not seem to be capturing a robust sense of compositionality. I will call this the \textbf{puzzle of compositional understanding}.

Empirical work shows that human learners do not, in fact, struggle with compositional interpretations in signaling-game settings. In iterated-learning experiments, participants quickly converge on signal systems whose parts can be recombined productively; receivers use these parts to infer the meanings of novel messages \citep{kirby2008}. Sender–receiver experiments paint the same picture: when roles alternate, both players reshape their codes so that individual signal elements map systematically onto meaning dimensions, and receivers generalize those mappings to unseen combinations \citep{Moreno2015}. Indeed, even young children show a bias to track function/content word patterns in artificial grammars in a way that supports later compositional interpretation, whereas adults rely on any salient regularity \citep{Nowak2016}. The puzzle I tackle therefore concerns model learners, not human ones. In contrast to the human results, standard Lewis-Skyrms reinforcement agents succeed at coordination yet fail to retain partial information carried by message components.\footnote{Some other related models on the evolution of language include \citet{kirby2000syntax, kirby2007evolution, nowak1999evolution, nowak2000evolution, batali1998computational}. }

In this paper, I argue that LaCroix and Franke are right: conventional signaling game models do not learn \emph{any} compositional signaling. However, the puzzle of compositional understanding can be solved.  I construct signaling game models in which the meaning of the signals is robust against replacement and forgetting, in precisely the way that Franke and LaCroix demand. Therefore, I argue that such models can in fact allow for the evolution of compositionality.

The models I present in this paper use a more sophisticated receiver, able to learn more complex patterns of information. I present two approaches, one minimalist and one generalist. In the minimalist approach, the receiver reinforces based on each atomic message that they receive. Formally, this kind of receiver could be interpreted as a simple feed-forward neural network. The generalist receiver reinforces on all the information available, in effect learning a full joint probability distribution.

Several approaches already exist in which agents do acquire a more robust kind of compositional communication, one that is resistant to the challenge posed by Franke and LaCroix. \citet{barrett2020compositional} provide three models of \emph{hierarchical composition games}, which augment conventional signaling games with executive agents, which can avoid the challenge raised by LaCroix and Franke. Under either an efficiency requirement or an explicit cost per signal, the executives steer learning so that each basic signal can come to encode just one semantic dimension. \citet{Franke2014} models agents equipped with a simple form of spill-over reinforcement learning, where successful updates diffuse to similar states and signals rather than remaining entirely local. This is enough for populations of otherwise unsophisticated learners to converge on stable, order-sensitive form–meaning mappings. Finally, \citet{steinert2020emergence} develops a model with sophisticated neural network senders and receivers in which compositional signaling evolves. In section \ref{sec:genuine}, I will suggest a reason why this model can avoid this puzzle of compositional understanding.

However, the two models presented here are distinct, and in some senses simpler than previous models. In both cases, the models solve the puzzle of compositional understanding by allowing for more sophisticated receivers, who explicitly consider the separate information considered in the atomic messages. The other features of the conventional signaling game models remain largely unchanged. These models also shed some light into why the puzzle of compositionality arises in standard models and I suggest some insights into the criteria that might be needed for compositional understanding more generally.

\section{Background}
\label{sec:sigames}

\subsection{Signaling Games}

In a typical signaling game, there are two players. One player (the sender) has access to the state of the world. The sender tries to convey information about the world to a second player (the receiver). Then the receiver performs an act, based only on the information that they have received.  Both players are rewarded if the receiver performs the right action corresponding to a given state of the world. However, the receiver does not know the meaning of the signals in advance: these meanings must be learned from rewards through successive iterations of the game.

More formally\footnote{For convenience, I will closely follow the formalism and notation of \citet{skyrms2010signals} and \citet{lacroix2023noncompositional} wherever possible, throughout this paper.}, a signaling game is defined by a tuple,

\begin{equation}
    \Sigma = \langle  S, M, A, P, \sigma, \rho, u \rangle,
\end{equation}

\noindent where $S = \{ s_0, \ldots s_k\}$ is a set of states of the world, $M = \{ m_0, \ldots, m_l \}$ is a set of possible messages that the sender can send and $A = \{a_0, \ldots a_n \}$ is a set of acts that the receiver can perform. Let $\Delta(X)$ be a set of probability distributions over a finite set, $X$. $P \in \Delta(S)$ assigns a probability distribution over the possible states of the world in $S$. We define the sender with a function from the states of the world to the messages that they send, $\sigma : S \rightarrow \Delta(M)$. Likewise, we define the receiver with a function from the messages that they receive to the actions that they can perform, $\rho : M \rightarrow \Delta(A)$. The players are rewarded through a function that assigns a utility based on whether performed actions appropriately match the state of the world $u : S \times A \rightarrow \mathbb{R}$.

For example, in the atomic 2-game, there are two equally probable states of the world, $s_0$ and $s_1$, two possible messages $m_0$ and $m_1$ and two possible actions, $a_0$ and $a_1$. Both players receive a reward of $1$ if act 0 is performed in state 0 or act 1 is performed in state 1. Otherwise, both players receive a reward of 0.

\begin{figure}
    \centering
    \begin{subfigure}{0.49\textwidth}
             \centering
	\begin{tikzpicture}[node distance={2cm}]
		\node (s0) {$s_0$};
		\node [below of=s0] (s1) {$s_1$};
		\node [right of=s0] (m0) {$m_0$};
		\node [right of=s1] (m1) {$m_1$};
		\node [right of=m0] (a0) {$a_0$};
		\node [right of=m1] (a1) {$a_1$};
		\draw [->] (s0) -- (m0);
		\draw [->] (s1) -- (m1);
		\draw [->] (m0) -- (a0);
		\draw [->] (m1) -- (a1);
	\end{tikzpicture}
    \end{subfigure}    
    \begin{subfigure}{0.49\textwidth}
                \centering
	\begin{tikzpicture}[node distance={2cm}]
		\node (s0) {$s_0$};
		\node [below of=s0] (s1) {$s_1$};
		\node [right of=s0] (m0) {$m_0$};
		\node [right of=s1] (m1) {$m_1$};
		\node [right of=m0] (a0) {$a_0$};
		\node [right of=m1] (a1) {$a_1$};
		\draw [->] (s0) -- (m1);
		\draw [->] (s1) -- (m0);
		\draw [->] (m0) -- (a1);
		\draw [->] (m1) -- (a0);
	\end{tikzpicture}
    \end{subfigure}
    \caption{The two signaling systems available in the atomic 2-game.}
    \label{fig:atomic2game}
\end{figure}
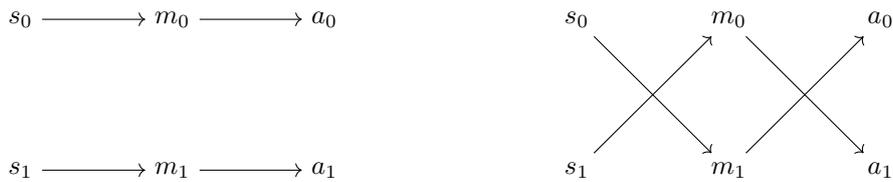

A signaling system arises if the sender and receiver coordinate strategies perfectly to maximize the payoff. Given that the meanings of the signals are initially arbitrary, there may be more than one possible signaling system in general. For example, in the atomic 2-game, there are two signaling systems (see figure \ref{fig:atomic2game}), one in which message 0 corresponds to state 0 and message 1 corresponds to state 1, \emph{or} one in which message 0 corresponds to state 1 and message 1 corresponds to state 0.

The sender and receiver learn to communicate through a process of reinforcement learning. The sender assigns weights to each (world, message) pair, and the receiver assigns weights to each (message, action) pair. The weights of the choice that each player has made are reinforced each turn according to the reward that they receive. A common choice is for each strategy to start with an equal reinforcement of 1, and for each strategy to be chosen with probability proportional to its accumulated reward \footnote{This is Richard Herrnstein’s matching law -- see \citealp{herrnstein1970law, roth1995learning, othmer1997aggregation, skyrms2000dynamic}.}, although other options are available. 

We can visualize this form of reinforcement learning with an urn model \footnote{For instance see \citep{polya1921uber, hoppe1984statistical}.}. For example, with the atomic 2-game, we can imagine that the sender has two urns, labeled $s_0$ and $s_1$, corresponding to each state of the world, with balls corresponding to messages $m_0$ and $m_1$. Meanwhile, the receiver has two urns, labeled $m_0$ and $m_1$ corresponding to the messages that they receive, with balls corresponding to acts $a_0$ and $a_1$. Players choose messages to send or acts to take by drawing a ball at random from the corresponding urn, with probability proportional to the number of such balls  in the urn. For example, suppose that the world is in state $s_0$, the sender chooses message $m_0$, and the receiver chooses act $a_0$, leading to the two players receiving a reward of $1$. Then the sender adds an $m_0$ ball to urn $s_0$, whilst the receiver adds an $a_0$ ball to urn $m_0$, making it more likely each player will opt for these strategies again in the future.

More precisely, at turn number $t$, if the world is in state $s_x$, then the sender's probability of picking any given message, $m_y$, is,

\begin{equation}
\sigma_t(m_y \mid s_x) = \frac{R_t(m_y \mid s_x) }{ \sum_i R(m_i \mid s_x)}, 
\label{eq:senderprobs}
\end{equation}

\noindent where $R_t(m_i \mid s_x)$ denotes the total reinforcement of message $m_i$ for a given state, $s_x$, (or the number of $m_i$ balls in the $s_x$ urn) at turn t. Likewise, given a message $m_y$, the receiver's probability of picking a given action $a_z$ is,

\begin{equation}
\rho_t(a_z \mid m_y) = \frac{R_t(a_x \mid m_y) }{ \sum_i R(a_i \mid m_y)}, 
\label{eq:receiverprobs}
\end{equation}

\noindent where $R_t(m_i \mid s_x)$ denotes the total reinforcement of act $a_i$ for a given message, $m_y$, (or the number of $a_i$ balls in the $m_y$ urn) at turn t.

\subsection{The information content of signals}
\label{sec:infocontent}
The entropy of a variable quantifies the amount of information one would need to learn its exact state \footnote{See \citealp{CoverThomas2006} for a more detailed overview.}. Let $X$ be a discrete variable with possible values $x_1, x_2, \ldots \in \mathcal{X}$. The information entropy of $X$ is defined as

\begin{flalign}
 H(X) = - \sum_{x \in \mathcal{X}} P(x) \text{log} P(x), 
\end{flalign}

\noindent where $P(x)$ is the probability of $X$ taking value $x$ \footnote{Where numerical values are specified, I will use logarithms of base 2 throughout his paper, but of course we can alternatively define information-theoretic quantities with any such choice.}. The entropy of a probability distribution is always greater than or equal to zero, $H(X) \geq 0$. In general, the information entropy of $X$ is maximal when all possible values are equally likely. If we gain information about the value of a random variable, then its information entropy decreases; if we know its exact value then the entropy will be 0. 

Suppose we have an agent with probabilities $P$. Following \cite{skyrms2010signals}, we can define the quantity of information that a signal $m_j$ conveys to the agent about a state $s_i$ of the world using their conditional credence about the state given that the signal was sent ($P(s_i \mid m_j$) and the unconditional, prior credence about the state, $P(s_i)$,

\begin{flalign}
 H(m_j, s_i) = \text{log} \frac{P(s_i \mid m_j)}{P(s_i)} .
\end{flalign}

\noindent This quantity is also known as the pointwise mutual information. However, in general, a signal can carry information about all states. Therefore, we can quantify the average \emph{total} information carried by a signal about the state of the world using a weighted sum of the information carried about each particular state,

\begin{flalign}
 H(m_j) = \sum_i P(s_i \mid m_j) \text{log} \frac{P(s_i \mid m_j)}{P(s_i)} .
 \label{eq:message_information}
\end{flalign}

\noindent This is the Kullback-Leibler divergence between the probability distribution conditional on the signal $m_j$ and the marginal probability distribution. It measures the total quantity information gained from moving from the marginal distribution to the conditional distribution, according to the conditional distribution.

For the agent, the expected information received per turn, will be given by the mutual information between the states ($S$) and the messages ($M$),

\begin{flalign}
I_\text{mut}(S;M) = \sum_{i,j} P(s_i, m_j) \log \frac{P(s_i \mid m_j)}{P(s_i)} ,
\end{flalign}

\noindent where $P(s_i, m_j)$ is the joint probability of state $s_i$ and message $m_j$. This quantity is always non-negative and symmetric ($I(S;M) = I(M;S)$), reaching zero only when the messages are wholly non-informative about the states.

The agent's beliefs about the frequency of messages may not be correct. Therefore, we will be interested in the true average information transmitted per turn. This will be given by,

\begin{flalign}
I_\text{average}(S;M) = \sum_{i,j} Q(m_j) P(s_i \mid m_j) \log \frac{P(s_i \mid m_j)}{P(s_i)} ,
 \label{eq:total_information}
\end{flalign}

\noindent where $Q$ represents the actual probabilities, with $Q(m_j)$ giving the actual probabilities that each message is sent. In other words, we weigh the information transmitted by each possible message by the probability that this message is sent.

We can also specify the information \emph{content} about specific states carried by a signal, not just its total quantity of information. If there are $k$ signals, then the information content can be represented by a $k$-dimensional vector,

\begin{alignat}{3}
I(m_j) = \left\langle \text{log} \frac{P(s_0 \mid m_j)}{P(s_0)}, \ldots, \text{log} \frac{P(s_{k-1} \mid m_j)}{P(s_{k-1})}\right\rangle,
\end{alignat}

\noindent where the $i$th element of the vector tells the information that signal $m_j$ provides about the corresponding state of the world. In general, there are two types of information we might be interested in: the information carried by the signals about the states of the world (the information interpreted by the sender), and the information carried by the signals about the acts (the information interpreted by the receiver) \footnote{The games considered in this paper will always have exactly one act optimal for each state of the world. Therefore, it will occasionally be convenient to talk about information about acts \emph{as if} it carried information about the state of the world directly. However, it will be very important to keep track of the information as interpreted by the sender and by the receiver.}.

It is convenient to represent the information carried by each signal about states or acts in tabular form. For example, consider the atomic 2-game. Suppose that the two signals provide no information about the state of the world (perhaps no learning has taken place, so the signals are entirely arbitrary). Then the information content of each signal about the states of the world would be

\begin{equation}
\begin{array}{@{}rrr@{}}
\toprule
       & s_0 & s_1 \\
\midrule
I(m_0) &  0 & 0  \\
I(m_1) &  0 & 0  \\
\bottomrule \\
\end{array}
\qquad
\begin{array}{@{}rrr@{}}
\toprule
       & a_0 & a_1 \\
\midrule
I(m_0) &  0 & 0  \\
I(m_1) &  0 & 0  \\
\bottomrule \\ 
\end{array}  
\begin{array}{r}
\\
 \\
 \\
.\\ 
\end{array} 
\end{equation}

\noindent Such a signal carries no information about the state of the world, or about the acts.

Alternatively, suppose that our agents have arrived at a signaling system, in which the receiver knows that $m_0$ always corresponds to $s_0$ and $m_1$ always corresponds to $s_1$. Then the information content would be,

\begin{equation}
\begin{array}{@{}rrr@{}}
\toprule
       & s_0 & s_1 \\
\midrule
I(m_0) &  1 & -\infty \\
I(m_1) &  -\infty & 1 \\
\bottomrule \\
\end{array}
\qquad
\begin{array}{@{}rrr@{}}
\toprule
       & a_0 & a_1 \\
\midrule
I(m_0) &  1 & -\infty \\
I(m_1) &  -\infty & 1 \\
\bottomrule \\
\end{array}
\begin{array}{r}
 \\
 \\
 \\
,\\ 
\end{array} 
\end{equation}

so each of the signals, $m_0$ and $m_1$, carries one bit of information, exactly enough to tell us the state of the world. Likewise, each signal exactly specifies the act that the receiver will take. The $-\infty$ elements tell us which state-signal or signal-act pairs have zero probability. \citet{skyrms2010signals} interprets this as a case of a signal carrying \emph{propositional information}. Thus the process of reinforcement learning breaks the symmetry between initially arbitrary signals, and information is created ``\emph{ex nihilo}'' (see \citealp[page 40]{skyrms2010signals}). 

\subsection{Compositionality}
\label{sec:compositionality}

A language is compositional if the meaning of each composite expression can be derived from the meaning of its parts and the way in which those parts are combined.\footnote{However, note that in natural languages, some morphemes may not bear information on their own but rather carry meaning only through their effect on other constituents of an expression, for instance \emph{only} in the phrase \emph{only child}.} For example, the meaning of this sentence I am now typing might derive from the meaning of these words and the way that I am combining them. Typically this is expressed as follows. Let $m_1, \ldots m_n$ be $n$ components of an expression, combined into a complex expression $\sigma(m_1, \ldots m_n)$, where $\sigma$ is some syntactic combination operation. Then if $I$ is an interpretation function, which assigns the meanings in this language,

\begin{equation}
    I \left( \sigma(m_1, \ldots m_n) \right) = f \left( I(m_1), \ldots I(m_n) \right),
\end{equation}

\noindent where $f$ is some function specific to the language in question. 

In the context of co-operative signaling games, it is most natural to interpret the meanings of expressions in terms of the information content conveyed by the signals, as we defined in section \ref{sec:infocontent}. The sender tries to send signals that best communicate their credences about the state of the world. And the meaning that the receiver interprets in the signals gives their credences about the state of the world. When a signaling system is reached, the sender and receiver agree about the meanings of expressions.

 However, in a compositional language, we should not need perfect transfer of information for \emph{some} information to be transmitted. Suppose that I know the meaning of $m_1$ but not $m_?$ and I receive the message $\sigma(m_1, m_?)$. Then this still conveys partial information about the possible values of $f \left( I(m_1), I(m_?) \right)$. For example, suppose that we are discussing the professor's choice of clothing today, and message $m_1$ conveys \emph{the clothing will be red}. \footnote{Note that some authors (e.g. \citealp{Franke2014}) maintain that conjunction is a rather trivial form of compositionality, because each conjunct already asserts its own content and the sentential meaning is just their joint assertion. Nevertheless, for convenience and simplicity, I will continue to treat conjunction as an example of compositionality.} Then the message $\sigma(m_1, m_?)$ provides me with partial information: I know that the clothing will be red, even if I am unclear about the item of clothing. Partial information has been conveyed by closing off some options: I know that the professor will not wear a chartreuse kimono or a fuchsia kaftan.  This capacity to convey partial information even when only some components are understood is a general feature of compositional languages.\footnote{For completeness, note that there are special cases in which partial expressions do convey precisely no information about the world. For instance, suppose I have a credence of exactly one half  that $m_?$ could mean an affirmation or negation. Then all I can discern from the complex expression $\sigma(m_1, m_?)$ is that the professor either wears red or does not wear red, with one half probability each. If I began with those same beliefs, then no information about the professor's clothing has been conveyed. However, if we shift my prior beliefs even slightly, or my credences about the meaning of $m_?$ even slightly, then at least \emph{some} partial information is conveyed. Note also that this is also true of signaling systems: if they only tell me what I already know, then no information about the world is conveyed.}

\subsection{Ostensibly Compositional Syntactic Games}
\label{sec:ostenisblycompositional}

The signals we have looked at so far have only carried atomic information. \citet{barrett2020compositional} propose ways of constructing \emph{syntactic games}, for which they argue the signals can carry compositional information (see also \citealp{barrett2006numerical, barrett2009coding, barrett2017selfassembling}). Let us look at one such example, a $4\times 4 \times 4$ two-sender signaling game. We will refer back to this game several times as a key example.

There are four states of the world, $s_0$, $s_1$, $s_2$, $s_3$, and four acts, $a_0$, $a_1$, $a_2$, $a_3$, with rewards of 1 if the act matches the world ($a_i = s_i)$ and 0 otherwise. Now, however, there are two senders, $\sigma_A$ and $\sigma_B$, each of whom can choose between two signals to send, $m^A_0$ or $m^A_1$ and $m^A_0$ or $m^A_1$ respectively. The one receiver must choose an action based on the information that they receive from both signals.

\begin{figure}
    \centering
	\begin{tikzpicture}[node distance={2cm}]
		\node (s0) {$s_0$};
		\node [below of=s0] (s1) {$s_1$};
  		\node [below of=s1] (s2) {$s_2$};
		\node [below of=s2] (s3) {$s_3$};
  
		\node [right=3cm of s0, text=red] (mA0) {$m^A_0$};
		\node [right=3cm of s1, text=red] (mA1) {$m^A_1$};
		\node [right=3cm of s2, text=blue] (mB0) {$m^B_0$};
		\node [right=3cm of s3, text=blue] (mB1) {$m^B_1$};
  
		\node [right=3cm of mA0] (a0) {$a_0$};
		\node [right=3cm of mA1] (a1) {$a_1$};
		\node [right=3cm of mB0] (a2) {$a_2$};
		\node [right=3cm of mB1] (a3) {$a_3$};
  
		\draw [->] (s0) -- (mA0);
		\draw [->] (s1) -- (mA0);
		\draw [->] (s2) -- (mA1);
		\draw [->] (s3) -- (mA1);
  
		\draw [->] (s0) -- (mB0);
		\draw [->] (s1) -- (mB1);
		\draw [->] (s2) -- (mB0);
		\draw [->] (s3) -- (mB1);

		\draw [->, red, out=0, in=180] (mA0) to (a0);
  		\draw [->, blue, out=0, in=180] (mB0) to (a0);
		\draw [->, red, out=0, in=180] (mA0) to (a1);
  		\draw [->, blue, out=0, in=180] (mB1) to (a1);
		\draw [->, red, out=0, in=180] (mA1) to (a2);
  		\draw [->, blue, out=0, in=180] (mB0) to (a2);
		\draw [->, red, out=0, in=180] (mA1) to (a3);
  		\draw [->, blue, out=0, in=180] (mB1) to (a3);

  \node [draw, fit=(mA0)(mA1), inner sep=8pt, label={[text=red, font=\itshape]above:sender A}] {};
  \node [draw, fit=(mB0)(mB1), inner sep=8pt, label={[text=blue, font=\itshape]below:sender B}] {};
            
	\end{tikzpicture}
    \caption{One possible signaling system available in the $4\times 4 \times 4$ two-sender signaling game. Each sender transmits a message if the world is in \emph{either} of the corresponding states. The receiver takes an action only when they receive \emph{both} the corresponding message from sender A (\textcolor{red}{red}) and the corresponding message from sender B (\textcolor{blue}{blue}).}
    \label{fig:compositional444game}
\end{figure}
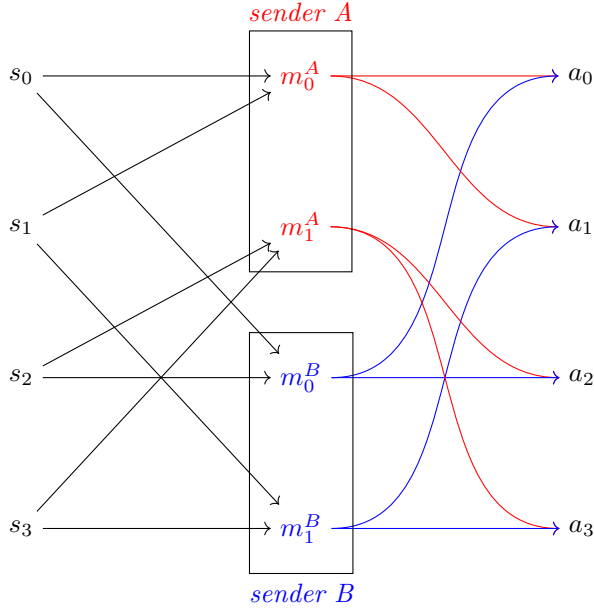

A signaling system requires that each pair of possible signals uniquely specifies one state of the world. One possible signaling system is shown in figure \ref{fig:compositional444game}. In this signaling system, each sender sends an incomplete message about the state of the world. Sender $A$ sends $m^A_0$ if the world is in state $s_0$ \emph{or} $s_1$, $m^A_1$ if the world is in state $s_2$ \emph{or} $s_3$, whereas sender $B$ sends $m^B_0$ if the world is in state $s_0$ \emph{or} $s_2$, $m^B_1$ if the world is in state $s_1$ \emph{or} $s_3$. However, the receiver's action is completely determined (and the corresponding state of the world is completely specified)  by the conjunction of the two messages: $a_0$ if they receive $m^A_0$ \emph{and} $m^B_0$, $a_1$ if they receive $m^A_0$ \emph{and} $m^B_1$, $a_2$ if they receive $m^A_1$ \emph{and} $m^B_0$, and $a_3$ if they receive $m^A_1$ \emph{and} $m^B_1$. 

A concrete example might help here. Suppose that the four states correspond to the professor's choice of clothing today: $a_0$ corresponds to a red dress, $a_1$ to a blue dress, $a_2$ to a red suit, and $a_3$ to a blue suit. Then the four messages have an obvious interpretation: $m^A_0$ corresponds to a dress, $m^A_1$ to a suit, $m^B_0$ to a red clothing, and $m^B_1$ to blue clothing. If the receiver correctly interprets these messages compositionally, then they should have learned that messages from sender A tell us the type of clothing and messages from sender B the color. 

We can see this by looking at the information content of the atomic messages about the states (i.e. the information as interpreted by the sender),

\begin{equation}
\begin{array}{@{}rrrrr@{}}
\toprule
       & s_0 & s_1 & s_2 & s_3 \\
\midrule
I(m^A_0) & 1 & 1 & -\infty & -\infty \\
I(m^A_1) & -\infty & -\infty & 1 & 1 \\
I(m^B_0) & 1 & -\infty & 1 & -\infty \\
I(m^B_1) & -\infty & 1 & -\infty & 1 \\
\bottomrule \\
\end{array}
\begin{array}{r}
\\
 \\
\\
 \\
 \\
.\\ 
\end{array} 
\label{eq:beforestates}
\end{equation}

\noindent On the other hand, if we look at the information carried possible conjunctions of the two signals about the acts (i.e. the information interpreted by the receiver), we see that the act is completely specified by the conjunctions of the signals,

\begin{equation}
\begin{array}{@{}rrrrr@{}}
\toprule
       & a_0 & a_1 & a_2 & a_3 \\
\midrule
I(m^A_0) \text{ and } I(m^B_0) & 2 & -\infty & -\infty & -\infty \\
I(m^A_0) \text{ and } I(m^B_1) & -\infty & 2 & -\infty & -\infty \\
I(m^A_1) \text{ and } I(m^B_0) & -\infty & -\infty & 2 & -\infty \\
I(m^A_1) \text{ and } I(m^B_1) & -\infty & -\infty & -\infty & 2 \\
\bottomrule \\
\end{array}
\begin{array}{r}
\\
 \\
\\
 \\
 \\
.\\ 
\end{array} 
\label{eq:beforeacts}
\end{equation}

At first glance, it might appear that the receiver is obtaining and interpreting compositional information -- in this case their action is determined by the conjunction of two atomic symbols. For example, if and only if the receiver sees messages $m^A_0$ \emph{and} $m^B_0$ they take action $a_0$, the appropriate action to receive a reward if the world is in state $s_0$. To return to our example, the receiver learns that the professor is wearing a red dress, from the two distinct messages, one telling him the clothing is a dress, the other telling him that the clothing is red. Analogous examples of apparently compositional communication have been built upon models of this essential form \citep{barrett2017selfassembling, barrett2020compositional}.

\section{The Puzzle of Compositional Understanding}
\label{sec:puzzle}

The signals sent by the sender in the $4\times 4 \times 4$ two-sender signaling game are compositional. However, upon closer inspection, we shall see that the signals are not being \emph{interpreted} compositionally by the receiver. To put it another way, the receiver is interpreting each conjunction of signals as a distinct atomic signal. Doubts along these lines have been raised by \citet{franke2016compositionality,  Franke2014, steinert2016compositional, lacroix2023noncompositional}.


\citet{lacroix2023noncompositional} demonstrates this with a simple example. Suppose that the $4\times 4 \times 4$ two-sender game has reached the signaling system described in section \ref{sec:ostenisblycompositional}. And let us tie this back to concrete example of professorial attire from section \ref{sec:ostenisblycompositional}. There are four atomic signals, $m^A_0$ (\emph{dress}), $m^A_1$ (\emph{suit}), $m^B_0$ (\emph{red}), and $m^B_1$ (\emph{blue}). The receiver knows how to interpret each compound signal, $m^A_0 \& m^B_0$ (\emph{red dress}), $m^A_0 \& m^B_1$ (\emph{blue dress}), $m^A_1 \& m^B_0$ (\emph{red suit}), and $m^A_1 \& m^B_1$ (\emph{blue suit}).

Now suppose that sender $B$ substitutes one of their signals, $m^B_0$ (\emph{red}) with an entirely new signal, $m^B_?$. This could be interpreted as an act of replacement of a signal, or forgetting the previous signal. Perhaps they have forgotten the English word \emph{red}, and instead substitute it with the French word \emph{rouge} to describe the color. So, they amend their behavior, so that any time they would have sent message $m^B_0$ (\emph{red}), they now send $m^B_?$ (\emph{rouge}).  In the urn analogy, they simply relabel all balls labeled with $m^B_0$ (\emph{red}) with balls labeled $m^B_?$ (\emph{rouge}) in the sender's urns. 

So now we have four different possible atomic signals, $m^A_0$ (\emph{dress}), $m^A_1$ (\emph{suit}), $m^B_?$ (\emph{rouge}), and $m^B_1$ (\emph{blue}).  However, let us suppose that our receiver does not understand any French. We will leave the receiver's balls unchanged, so they start with zero balls labeled $m^B_?$ (\emph{rouge}). Now if the senders deliver the message $m^A_0 \& m^B_?$ (\emph{dress} and \emph{rouge}) to the receiver, how does this change the information content of the signal?

The inclusion of a novel signal does not meaningfully change the informational content available to the sender about the states of the world: except for the relabelling their reinforcement remains the same. So except for the exact substitution of the message $m^B_0$ (\emph{red}) with $m^B_?$ (\emph{rouge}), the information content of the signals will be identical to that in equation \ref{eq:beforestates},

\begin{equation}
\begin{array}{@{}rrrrr@{}}
\toprule
       & s_0 & s_1 & s_2 & s_3 \\
\midrule
I(m^A_0) & 1 & 1 & -\infty & -\infty \\
I(m^A_1) & -\infty & -\infty & 1 & 1 \\
I(m^B_?) & 1 & -\infty & 1 & -\infty \\
I(m^B_1) & -\infty & 1 & -\infty & 1 \\ 
\bottomrule \\ 
\end{array} 
\begin{array}{r}
\\
 \\
\\
 \\
 \\
.\\ 
\end{array} 
\label{eq:afterstates}
\end{equation}

\noindent However, the information available to the receiver about the acts has changed! After all, the receiver has not yet learned the meaning of the new message $m^B_?$ (they do not understand the French word \emph{rouge}, and this new signal starts with zero reinforcement). Therefore, at least some information must have been lost, when compared to equation \ref{eq:beforeacts}. The information about the acts is now,

\begin{equation}
\begin{array}{@{}rrrrr@{}}
\toprule
       & a_0 & a_1 & a_2 & a_3 \\
\midrule
I(m^A_0) \text{ and } I(m^B_?) & 0 & 0 & 0 & 0 \\
I(m^A_0) \text{ and } I(m^B_1) & -\infty & 2 & -\infty & -\infty \\
I(m^A_1) \text{ and } I(m^B_?) & 0 & 0 & 0 & 0 \\
I(m^A_1) \text{ and } I(m^B_1) & -\infty & -\infty & -\infty & 2 \\
\bottomrule \\
\end{array}
\begin{array}{r}
\\
 \\
\\
 \\
 \\
,\\ 
\end{array} 
\label{eq:afteracts}
\end{equation}

\noindent where we interpret the signals as \emph{dress and rouge}, \emph{dress and blue}, \emph{suit and rouge}, and \emph{suit and blue} respectively.

\begin{figure}
    \centering
    \includegraphics[width=0.7\linewidth]{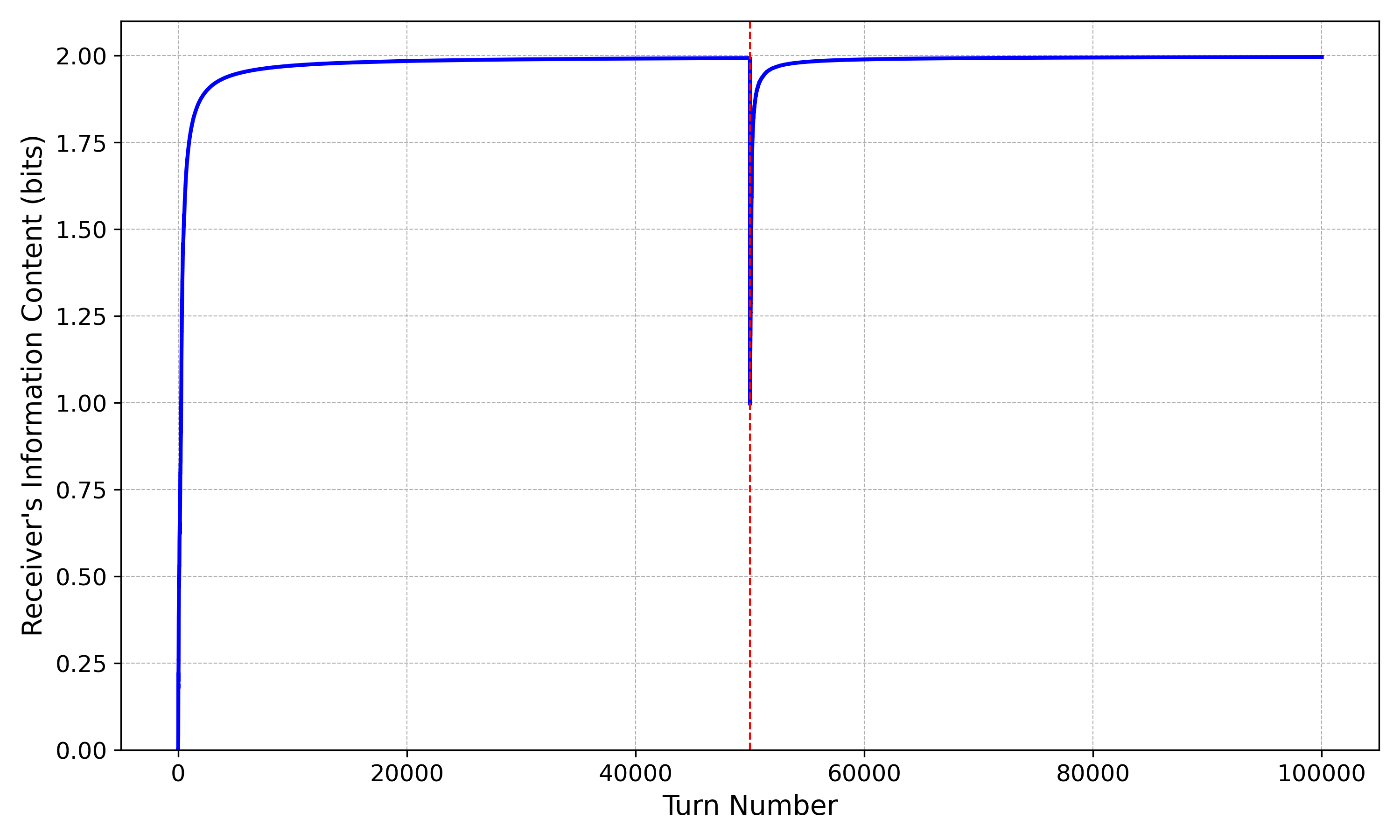}
    \caption{Simulation of the receiver's average information in the conventional $4\times4\times4$ two-sender signaling game, with the signal replaced at turn 50,000 (red, dashed line). Almost one bit of information is lost.}
    \label{fig:simulation_4x4x4_receiver_information_content}
\end{figure}

The problem is that rather more information has been lost than we would expect if the receiver were interpreting these signals compositionally. Observe that the sender cannot glean any meaning from any signal containing the new signal, \emph{rouge}: the information content is zero for every column in those two rows. That is, all of the information from any of the compositional signals involving $m^B_0$ (\emph{red}) has been wiped out altogether, whilst all signals containing the new message $m^B_?$ (\emph{rouge}) convey no information. To put it another way, messages $m^A_0$ (\emph{dress}) or $m^A_1$ (\emph{suit})will initially convey \emph{no information} to the receiver when they are sent in conjunction with the new signal $m^B_?$ (\emph{rouge}). In total, an average of one bit of information has been lost, according to equation \ref{eq:total_information}, as shown in figure \ref{fig:simulation_4x4x4_receiver_information_content}.

If the signals \emph{were} being interpreted compositionally, then $m^A_0$ and $m^A_1$ should still convey some information, regardless of the substitution. For instance, $m^A_0$ (\emph{dress}) should still individually correspond to the world being in one of two states, $s_0$ or $s_1$ (\emph{red dress} or \emph{blue dress}); likewise $m^A_1$ should correspond to the world being in state $s_2$ or $s_3$ (\emph{red suit} or \emph{blue suit}). The addition of the unknown signal, $m^B_?$ (\emph{rouge}) would not be able to eradicate the known meaning of these messages. By equation \ref{eq:total_information}, we would expect only one half of a bit of information to be lost on average. The total information would be,

\begin{equation}
\begin{array}{@{}rrrrr@{}}
\toprule
       & a_0 & a_1 & a_2 & a_3 \\
\midrule
I(m^A_0) \text{ and } I(m^B_?) & 1 & 1 & -\infty & -\infty \\
I(m^A_0) \text{ and } I(m^B_1) & -\infty & 2 & -\infty & -\infty \\
I(m^A_1) \text{ and } I(m^B_?) & -\infty & -\infty & 1 & 1 \\
I(m^A_1) \text{ and } I(m^B_1) & -\infty & -\infty & -\infty & 2 \\
\bottomrule \\ 
\end{array}
\begin{array}{r}
\\
 \\
\\
 \\
 \\
.\\ 
\end{array} 
\label{eq:afteractscompos}
\end{equation}

Recall from \ref{sec:compositionality}: that the components of a compositional signal should each communicate \emph{some} partial information. If the receiver were truly interpreting these messages compositionally, then they should have learned that messages from sender A tell us the type of garment and messages from sender B the color. Then, even though they do not know the meaning of $m^B_?$ (\emph{rouge}), the conjunction of $M^A_0$ (\emph{dress}) and $m^B_?$ (\emph{rouge}) should still convey partial information: the professor is wearing a dress, of unknown color. The replacement of the word \emph{red} should not prevent the receiver from understanding the word \emph{dress}.

Thus, although the senders transmit compositional signals, it seems that the receiver must be interpreting each of the four pairs of signals atomically. That is, they understand a message like $m^A_0 \& m^B_?$ as a single unit, rather than understanding the meaning from the meaning of its parts and the way that they are combined. This general argument applies to the ostensibly compositional signaling game models of \citet{barrett2006numerical, barrett2017selfassembling, franke2016compositionality, scott2012communication}. \citet{lacroix2023noncompositional} draws a general conclusion from this lesson:  the problem is with the syntactic composition of atomic messages. Perhaps ``focusing exclusively on syntax in discussing the evolution of compositionality under the signaling-game framework is misguided'' \citep[p.~12]{lacroix2023noncompositional}. However, if we could construct a syntactic model in which the receiver \emph{does} interpret the signals compositionally, then we need not succumb to this pessimistic conclusion.

\section{Genuine Compositionality}
\label{sec:genuine}

In retrospect, it is not surprising that the receivers were not able to interpret these signals compositionally in the $4 \times 4 \times 4$ two-sender signaling game. The problem lies with the information that we allowed the receiver to learn. Recall that the receiver was only storing information about \emph{complete} pair of signals, without storing any information about its components. Recalling equation \ref{eq:receiverprobs}, the receiver's probabilities of choosing an action at each time step were restricted to probabilities of each act conditional on the pairs of signals,

\begin{equation}
\rho_t(a_z \mid m^A_x \& m^B_y) = \frac{R_t(a_z \mid m^A_x \& m^B_y) }{ \sum_i \sum_j R(a_i \mid m^A_x \& m^B_y)}, 
\label{eq:444receiverprobs}
\end{equation}

\noindent where $R_t(a_i \mid m^A_x \& m^B_y)$ is the total reinforcement of act $a_i$ given the conjunction of both message $m^A_x$ \emph{and} message $m^B_y$. The receiver could only reinforce strategies of this form. Or, in the urn analogy, the receiver has only four urns: one for each pair, $m^A_0$ and $m^B_0$, $m^A_0$ and $m^B_1$, $m^A_1$ and $m^B_0$ and $m^A_1$ and $m^B_1$. There is no reinforcement on responses to the signal components. As such, the receiver is restricted to necessarily treat each message pair as basic, rather than its atomic components.

How might we loosen this restriction? \citet{franke2016compositionality} proposes using a method of spill-over reinforcement learning. In this process, when a particular signal-action pair is reinforced, other  signal-action pairs are also reinforced to a lesser degree according to a similarity metric.  This innovation allows agents to generalize from learned associations to novel situations. But it requires us to install a similarity metric into the agents a priori, and these similarity metrics that do much of the heavy-lifting. In a more naturalistic model, we would like the agents should spontaneously learn the similarities between the components. \citet{steinert2016compositional} questions whether these models have a genuine syntactic structure that gets compositionally interpreted.

I propose two plausible approaches. A simple approach would be to allow the receiver to learn probabilities conditional on the atomic messages from each sender-- let us call this the \textbf{minimalist approach}. A bolder approach might be to relax the restriction fully, and let the receiver learn about all of the message components and their combinations that he receives, as if they are learning the full joint probability distribution -- let us call this the \textbf{generalist approach}. Either of these approaches will work, but both are worth considering.

\subsection{The Minimalist Approach}

In the minimalist approach, we reinforce acts conditional on each of the atomic messages that they receive, rather than the combination of messages. In the urn analogy, the receiver would have four urns, one for each of the input messages, $m^A_0$,  $m^A_1$, $m^B_0$ and $m^B_1$. Suppose that the act $a_0$ is to be reinforced upon receiving messages $m^A_0$ and $m^B_0$: then an $a_0$ ball would be placed in both the $m^A_0$ and the $m^B_0$ urns. Using our clothing analogy, we replace the urns representing \emph{red dress}, \emph{blue dress}, \emph{red suit} and \emph{blue suit} with urns representing \emph{red}, \emph{blue}, \emph{dress} and \emph{suit}.

Suppose that the information content carried by the signals about the states is exactly the same as before. Then this would allow for a signaling system in which the information carried by the messages about the acts corresponds exactly to the information about the states, for example,

\begin{equation}
\begin{array}{@{}rrrrr@{}}
\toprule
       & a_0 & a_1 & a_2 & a_3 \\
\midrule
I(m^A_0) & 1 & 1 & -\infty & -\infty \\
I(m^A_1) & -\infty & -\infty & 1 & 1 \\
I(m^B_0) & 1 & -\infty & 1 & -\infty \\
I(m^B_1) & -\infty & 1 & -\infty & 1 \\
\bottomrule \\
\end{array}
\begin{array}{r}
\\
 \\
\\
 \\
 \\
.\\ 
\end{array} 
\label{eq:minimalistacts}
\end{equation}
\noindent Clearly, if the message $m^B_0$ were replaced by $m^B_?$, only the information pertaining to $m^B_0$ would be lost. The other message would give partial information,
\begin{equation}
\begin{array}{@{}rrrrr@{}}
\toprule
       & a_0 & a_1 & a_2 & a_3 \\
\midrule
I(m^A_0) & 1 & 1 & -\infty & -\infty \\
I(m^A_1) & -\infty & -\infty & 1 & 1 \\
I(m^B_?) & 0 & 0 & 0 & 0 \\
I(m^B_1) & -\infty & 1 & -\infty & 1 \\
\bottomrule \\
\end{array}
\begin{array}{r}
\\
 \\
\\
 \\
 \\
.\\ 
\end{array} 
\label{eq:minimalistactspartial}
\end{equation}

So in such a system, the receiver does seem to interpret the signals in a way that is genuinely compositional. Each sender's message conveys partial information to the receiver, and only by combining the messages does the receiver acquire the full information. If a sender's message is replaced, \emph{no} other information is lost.

However, we still need to show that the sender and receiver could arrive at this signaling system in the first place. The key here is how the receiver should  pick an act, conditional on both messages that they receive. Naively, one might simply pick the act  proportional to the weighted sum of a strategy's reinforcement given each message, in the same way as the previous signaling games,

\begin{equation}
\rho^\text{naive}_t(a_z \mid m^A_x \& m^B_y) =
\frac{R_t(a_z \mid m^A_x) + R_t(a_z \mid m^B_y) }
{ \sum_i R_t(a_i \mid m^A_x) + R_t(a_i \mid m^B_y)}. 
\label{eq:naiveminimalistreceiverprobs}
\end{equation}

\noindent However, it is easy to see that this would not lead to this signaling system. Consider the receiver's long-run reinforcements in each of their four urns, $m^A_0$,  $m^A_1$, $m^B_0$ and $m^B_1$. In the long run, we would expect the receiver's urn $m^A_0$ to receive roughly equal reinforcement with the balls for acts $a_0$ and $a_1$, and the receiver's urn $m^B_0$ to have roughly equal reinforcement for acts $a_0$ and $a_2$. Now suppose that the senders transmit messages $m^A_0$ and $m^B_0$: the receiver would pick the correct corresponding act $a_0$ around half the time, but they would be just as likely to choose an incorrect act, $a_1$ or $a_2$ each about one quarter of the time .

To put it another way, upon receiving the component messages \emph{red} and \emph{dress}, three possible acts would all have non-trivial reinforcement, corresponding to \emph{red dress}, \emph{red suit} and \emph{blue dress}. Only the act corresponding to the \emph{blue suit} would never be reinforced by the messages \emph{red} and \emph{dress}. So if the acts are chosen in direct proportion to their reinforcement, the receiver would often choose the wrong act.

But nothing about the structure of the signaling game compels us to require the receiver to choose acts in this way. They do not have to pick acts with probabilities directly proportional their total reinforcement. An alternative would be for the receiver to instead pick strategies using some \emph{activation function}, $f$, chosen so as to ensure the choices are more sharply peaked around the correct action,

\begin{equation}
\rho^\text{minimalist}t(a_z \mid m^A_x \& m^B_y) = f(\rho^\text{naive}_t(a_z \mid m^A_x \& m^B_y)).
\end{equation}

\begin{figure}
    \centering
    \includegraphics[width=0.7\linewidth]{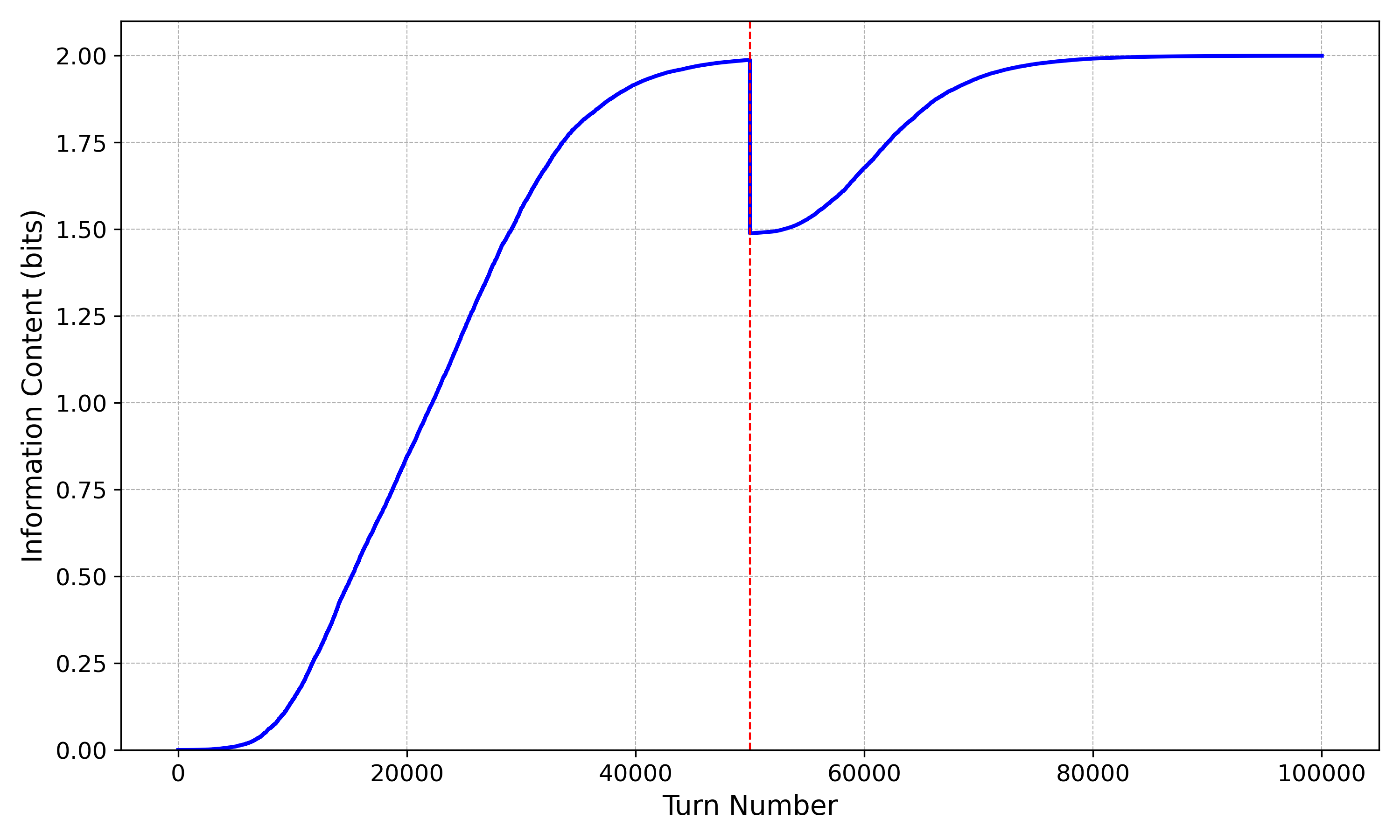}
    \caption{Simulation of the tempered-softmax minimalist receiver's average information in the $4\times4\times4$ two-sender signaling game, with the signal replaced at turn 50,000 (red, dashed line). Almost half a bit of information is lost. For visibility purposes, the learning rate is slowed with a high temperature, $T = 2000$.}
    \label{fig:minimalist_4x4x4_information_content}
\end{figure}

\noindent A sigmoid shape would be especially useful here, making the difference between strongly and weakly reinforced actions becomes more pronounced. This means that when receiving messages \emph{red} and \emph{dress}, the receiver would be much more likely to choose the strongly reinforced \emph{red dress} over the partly reinforced \emph{red suit} or \emph{blue dress}. One possible choice is the tempered softmax function,

\begin{equation}
f_\text{TSM}(x_j) = \frac{\exp(x_j/T)}{\sum_i \exp(x_i/T)},
\end{equation}

\noindent where the \emph{temperature}, $T$, controls the steepness of the function. With this in place, the sender and receiver can efficiently arrive at a signaling system. When replacement takes place, only a half a bit of information is lost, as we would expect for compositional signaling (see figure \ref{fig:minimalist_4x4x4_information_content}).

\begin{figure}
    \centering
    \begin{tikzpicture}[x=1.5cm, y=1.5cm, >=stealth, node distance=2cm, every node/.style={minimum size=1cm}]

\node[circle] (l1) at (0,0.6) {Inputs};
\node[circle] (l2) at (1.5,0.6) {Weights};
\node[circle] (l3) at (3,0.6) {Sum};
\node[circle] (l3) at (5,0.6) {Activation function};
\node[circle] (l3) at (7,0.6) {Output};

\node[circle, draw, thick] (mA0) at (0,0) {$m^A_0$};
\node[circle, draw, thick, below of=mA0] (mA1) {$m^A_1$};
\node[circle, draw, thick, below of=mA1] (mB0) {$m^B_0$};
\node[circle, draw, thick, below of=mB0] (mB1) {$m^B_1$};

\node[right of=mA0] (wA0) {$R^A_0$};
\node[right of=mA1] (wA1) {$R^A_1$};
\node[right of=mB0] (wB0) {$R^B_0$};
\node[right of=mB1] (wB1) {$R^B_1$};

\node[rectangle, draw, thick, right of=wA1, xshift=0.5cm] (sum) at (1.5,-2) {$\rho^\text{naive}$};

\node[rectangle, draw, thick] (f) at (5,-2){$f(\rho^\text{naive})$};

\node[circle, draw, thick] (output) at (6.5,-2) {action};

\draw[->,>=stealth,thick] (mA0) -- (wA0) -- (sum);
\draw[->,>=stealth,thick] (mA1) -- (wA1) -- (sum);
\draw[->,>=stealth,thick] (mB0) -- (wB0) -- (sum);
\draw[->,>=stealth,thick] (mB1) -- (wB1) -- (sum);

\draw[->,>=stealth,thick] (sum) -- (f) -- (output);
                    
        \end{tikzpicture} 
        \caption{The minimalist receiver in the $4\times 4 \times 4$ two-sender signaling game, interpreted as a simple perceptron.}
\label{fig:neuralnet}
\end{figure}
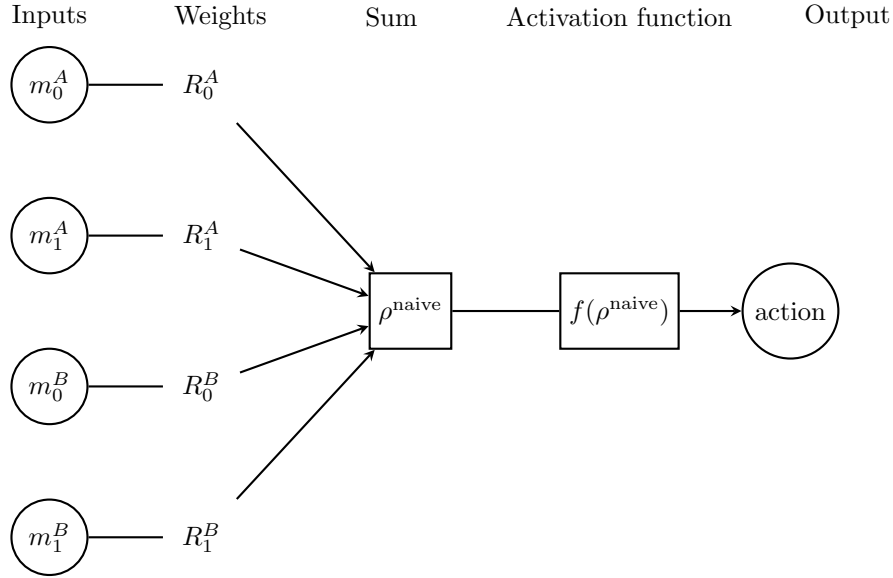

In effect, we could interpret such a minimalist receiver as a very simple, feed-forward perceptron \emph{neural network}. In this interpretation, the four atomic messages forming the input layer, the four acts as the output layer, the function $\rho^\text{naive}$ as the weights to be learned, and the $f$ as the activation function, as depicted in figure \ref{fig:neuralnet}. In retrospect, this should not be a surprise: neural networks are designed precisely to learn composite information.\footnote{Indeed, the historical controversy over the ability of neural networks to learn compositional information seems to closely parallel the contemporary debate about compositionality in signaling games (see \citealp{minsky1972, Goodfellow-neuralnets, olazaran1996, werbos1974beyond, rumelhart1986learning}). Indeed, these observations also seem to complement \citep{steinert2020emergence}, who makes explicit use of neural network agents to learn complex compositional information in signaling games. }

One might wonder whether the introduction of such an activation function is in some way ad hoc? Perhaps, although it should be noted that the conventional choice of choosing a strategy in direct proportion to the total reinforcement would be also arbitrary, in the absence of any particular justification. \footnote{Potentially, in an even more general setting, one could allow a good functional form of $f$ to be learned as well.} After all, we should also think of equation \ref{eq:receiverprobs} as specifying some choice of activation.

\subsection{The Generalist Approach}

In the generalist approach, we allow the receiver to effectively learn about all of the message components and their combinations (we can think of this as if they are learning a full joint probability distribution), that is $\rho(a_0, a_1, a_2, a_3, m^A_0, m^A_1, m^B_0, m^B_1)$. This could be implemented in a number of ways.

One possible implementation using the urn analogy is as follows. We allow the receiver to have one-variable urns, corresponding to the individual messages and acts ($m^A_0$, $a_0$, etc.); two-variable urns, corresponding to the message pairs, or message-act pairs ($m^A_0 \text{ and } m^B_0$, $m^A_0 \text{ and } a_0$, etc.), up to the maximum number of possible variables: in this case three (two messages and one act). Urns corresponding only to messages (and combinations of messages) are reinforced whenever that message (or combination) is received. For example, upon receiving the message combination $m^A_0 \text{ and } m^B_0$, a ball in placed in three urns: the one-variable urns $m^A_0$ and $m^B_0$, and the two-variable urn, $m^A_0 \text{ and } m^B_0$. These urns correspond to the receiver learning about how likely different combinations of messages are to arrive together. When an act is successfully rewarded, a ball is placed in the urns corresponding to any part of that message and the act. For example, if the receiver is rewarded for choosing $a_0$ in response to the previous message, a ball is placed in three urns: the two-variable urns $m^A_0 \text{ and } a_0$ and $m^B_0 \text{ and } a_0$, and the three-variable urn, $m^A_0 \text{ and } m^B_0 \text{ and } a_0$. These urns correspond to the receiver learning which acts receive reward in response to different combinations of messages.

In other words, our receiver learns on the basis of \emph{both} the full message, and its components (as well as learning of any correlations between the messages themselves!). Our receiver learns about which acts are reinforced on the basis of a message containing \emph{red}, a message containing \emph{dress} \textbf{and} on the full message being \emph{red dress}.

We might imagine them piecing together a full joint probability distribution about all these components. \footnote{Thinking of this in terms of probabilities is a convenience. However the model is still purely syntactic, in the sense of \citet{barrett2007conventionality, barrett2009coding}.} Let $M^n_x$ stand for an arbitrary combination of $n$ messages. Then we might imagine the receiver's reinforcements as them learning unconditional probability distributions about the different messages and message components, and about which acts receive reinforcement based on the different messages and message components,

\begin{align}
    \rho^\text{generalist}_t(M^n_x) &= \frac{R_t(M^n_x)}{\sum_i R_t(M^n_i)}  \label{eq:implicitlearnedprobs} \\
    \rho^\text{generalist}_t(a_y, M^n_x) &= \frac{R_t(a_y, M^n_x)}{\sum_i \sum_j R_t(a_j, M^n_i)} \label{eq:implicitlearnedrewards}.
\end{align}

\noindent We could think of equation \ref{eq:implicitlearnedprobs} corresponds to the implicit learned probabilities about combinations of messages, whilst equation \ref{eq:implicitlearnedrewards} corresponds to the implicit learned probabilities about which acts receive rewards for different combinations of messages. With this in place, we then set the receiver to choose acts according to the implicit conditional probabilities of the act given the \emph{full message} that they have received, i.e.,

\begin{align}
    \rho^\text{generalist}_t (a_y \mid M^n_x) =  \frac{ \rho(a_y, M^n_x)}{ \sum_j{ \rho(a_j, M^n_x)}} .
\end{align}


\noindent Finally, we need to specify a rule for the appropriate initial reinforcement if a new message is introduced at some turn, $T$, such as $m^B_?$. Observe that adding a new message must involve supplementing the system with not just one new urn, but many new urns, for each possible message and message-act combination involving $m^B_?$. This could be done in a number of different ways. One tempting choice would be to simply add all new messages with a reinforcement of 1,

\begin{align}
    R_T(m^B_? \text{ and } M^n_i) &= 1,
\end{align}

\noindent for all message combinations, $M^n_i$. Such a choice might be appropriate in many settings; however it would destroy some of the informational content of the existing messages. After all, then the probability of any given act, conditional on a signal, $m^B_? \text{ and } M^n_x$,  that includes the new message would be,

\begin{align}
    \rho^\text{information-erasing}_T (a_y \mid m^B_? \text{ and } M^n_x) = \frac{ \rho(a_y, m^B_?, M^n_x)}{ \sum_j \rho(a_j, m^B_?, M^n_x)} = \frac{1}{N_a} ,
\end{align}

\noindent where $N_a$ is the number of acts available, independent of the rest of the signal, $M^n_x$. If we view all new messages involving the new signal to carry the same information, then we effectively no longer treat those other messages as carriers of compositional information. Some information from the other signals $M^n_x$ will be erased, just as in the examples in section \ref{sec:ostenisblycompositional}. In the $4\times4\times4$ two-sender game, the information carried by the messages would be,

\begin{equation}
\begin{array}{@{}rrrrr@{}}
\toprule
       & a_0 & a_1 & a_2 & a_3 \\
\midrule
I(m^A_0) \text{ and } I(m^B_?) & 0 & 0 & 0 & 0 \\
I(m^A_0) \text{ and } I(m^B_1) & -\infty & 2 & -\infty & -\infty \\
I(m^A_1) \text{ and } I(m^B_?) & 0 & 0 & 0 & 0 \\
I(m^A_1) \text{ and } I(m^B_1) & -\infty & -\infty & -\infty & 2 \\
\bottomrule \\
\end{array}
\begin{array}{r}
\\
 \\
\\
 \\
 \\
.\\ 
\end{array} 
\label{eq:afteractsgeneralist}
\end{equation}

Instead, in order to preserve the informational content of existing messages, the receiver should begin with the assumption that the existing messages are probabilistically independent of the new message, 

\begin{align}
    \rho^\text{information-preserving}_T(M^n_i \mid m^B_?) &= \rho(M^n_i), \\ 
    \rho^\text{information-preserving}_T(a_j \text{ and } M^n_i \mid m^B_?) &= \rho(a_j \text{ and } M^n_i),
\end{align}
\noindent for all messages $M^n_i$ and acts, $a_j$. Then, the probability of any given act, conditional on a signal that includes the new message would be,

\begin{align}
    \rho^\text{information-preserving}_T (a_y \mid m^B_? \text{ and } M^n_x) &= \frac{ \rho(a_y, m^B_?, M^n_x)}{ \sum_j \rho(a_j, m^B_?, M^n_x)} \\
    &= \frac{ \rho(a_y, M^n_x)}{ \sum_j \rho(a_j, M^n_x)} ,
\end{align}

\noindent exactly preserving the informational content of the other messages in the signal, $M^n_x$. The message of each signal in the $4\times4\times4$ 2-messenger game would then be exactly as one would expect if the signals are being interpreted compositionally,

\begin{equation}
\begin{array}{@{}rrrrr@{}}
\toprule
       & a_0 & a_1 & a_2 & a_3 \\
\midrule
I(m^A_0) \text{ and } I(m^B_?) & 1 & 1 & -\infty & -\infty \\
I(m^A_0) \text{ and } I(m^B_1) & -\infty & 2 & -\infty & -\infty \\
I(m^A_1) \text{ and } I(m^B_?) & -\infty & -\infty & 1 & 1 \\
I(m^A_1) \text{ and } I(m^B_1) & -\infty & -\infty & -\infty & 2 \\
\bottomrule \\ 
\end{array}
\begin{array}{r}
\\
 \\
\\
 \\
 \\
.\\ 
\end{array} 
\label{eq:afteractscomposgeneralist}
\end{equation}

\begin{figure}
    \centering
    \includegraphics[width=0.7\linewidth]{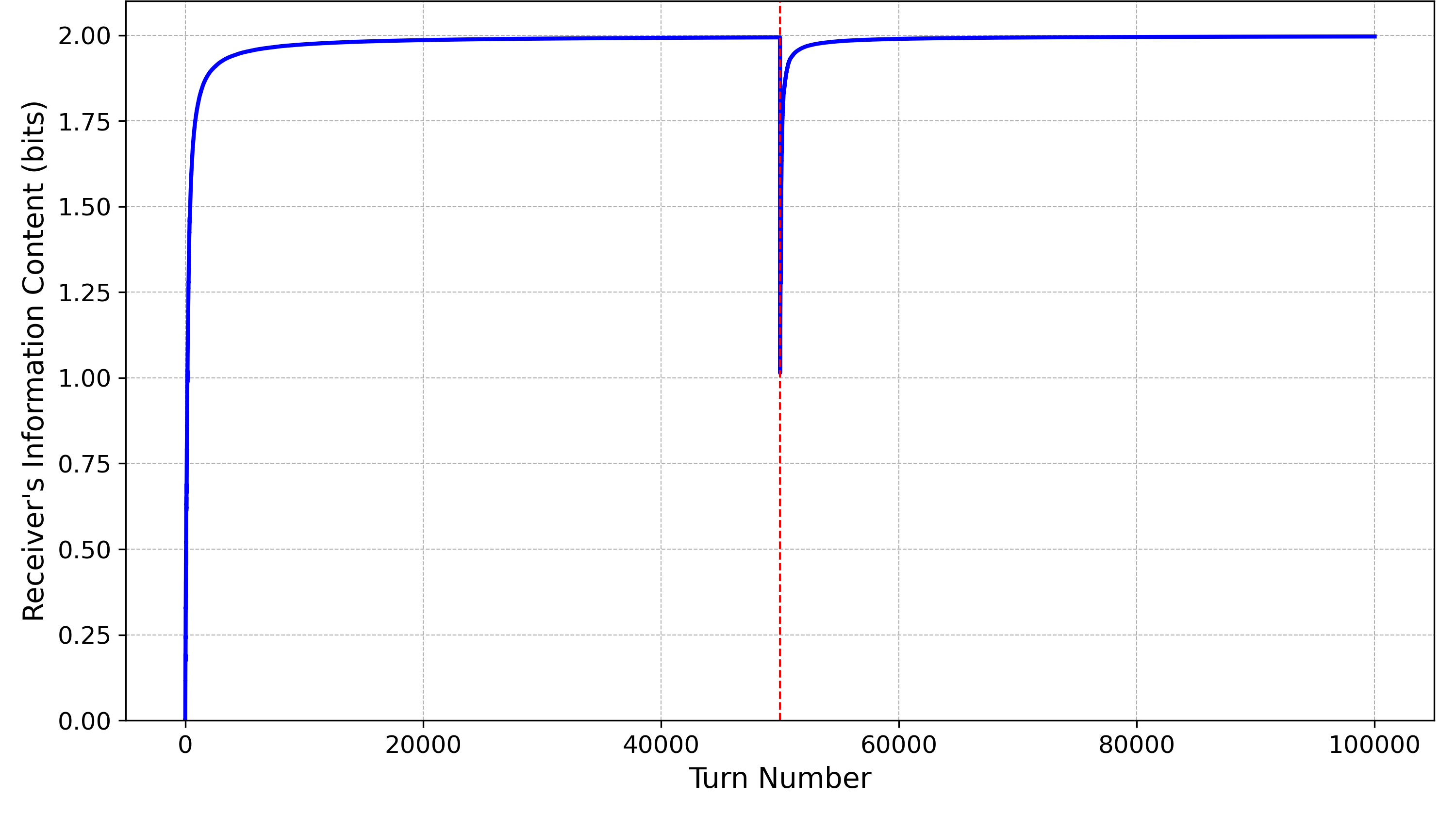}
    \caption{Simulation of the information-erasing generalist receiver's average information in the $4\times4\times4$ two-sender signaling game, with the signal replaced at turn 50,000 (red, dashed line). Almost one bit of information is lost.}
    \label{fig:generalist_4x4x4_information_content_erasing}
\end{figure}

\begin{figure}
    \centering
    \includegraphics[width=0.7\linewidth]{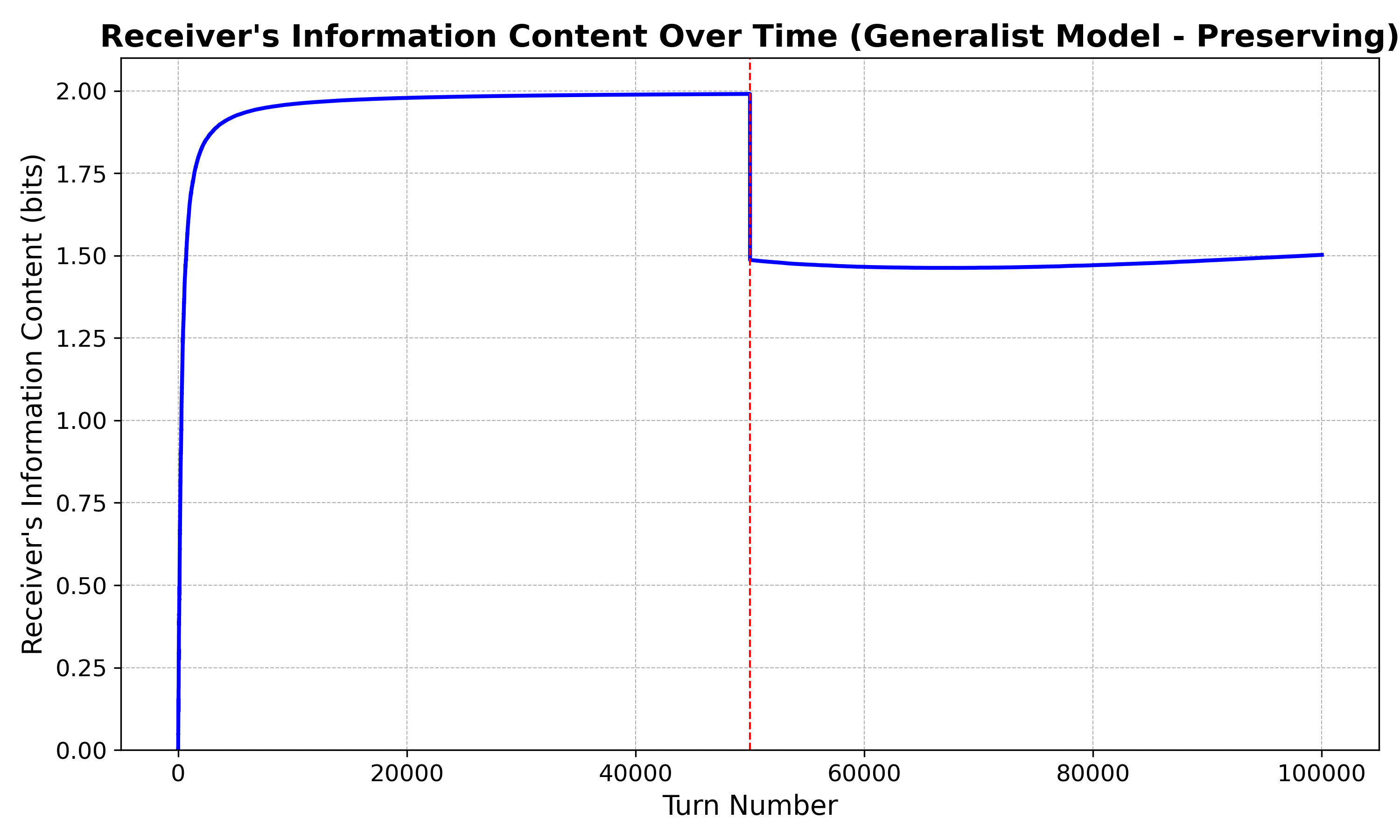}
    \caption{Simulation of the information-preserving generalist receiver's average information in the $4\times4\times4$ two-sender signaling game, with the signal replaced at turn 50,000 (red, dashed line). Almost half a bit of information is lost.}
    \label{fig:generalist_4x4x4_information_content_preserving}
\end{figure}

\noindent Simulation results are shown in figures \ref{fig:generalist_4x4x4_information_content_erasing} and \ref{fig:generalist_4x4x4_information_content_preserving}. Observe that the information-preserving generalist loses less information but also recovers more slowly after the message is erased. This is a feature of the urn-reinforcement learning: having retained much of the information about the existing signals, the urns remain significantly reinforced even after the old message is erased. This is a feature of the urn-model of reinforcement learning, and would not necessarily hold with other kinds.


Consider again our professorial vestments example. We can think of the two ways of incorporating a new message as follows. Suppose that the message \emph{red} has been erased and the receiver obtains the mysterious new messages \emph{rouge} and \emph{dress}. On the information-erasing approach, they regard this as a truly new message with no prior reinforcement. As such, this message erases prior information that they had learned about the meaning of the word \emph{dress} in this context. On the information-preserving approach, they could assume that the message contained in this new message \emph{rouge} is statistically independent of what they already knew about the word \emph{dress}. As such, they believe that the composite message \emph{rouge dress} retains all of their previous knowledge and so tells them that the garment being worn is at least a dress, regardless of its color.

One advantage of this generalist approach is that grants us greater flexibility. We can choose to model a new signal as either information-destroying or information-preserving, depending on what we consider appropriate in another given situation. Either of these approaches might be well-suited to modeling the integration of different kinds of new information. Another advantage is that the generalist approach does not depend on an arbitrary activation function as in the minimalist approach. One final advantage is that it presents a far more general learning framework than the minimalist approach: we allow the receiver to learn any part of the joint probability distribution over all the variables under consideration.~\footnote{Where one could interpret the minimalist receiver as a simple feed-forward neural network, one might interpret the receiver in the generalist procedure as implicitly learning the probabilities in a Bayesian network.}

Another advantage of the generalist model over the minimalist model is that it reaches a signaling system just as often as the traditional model. The reason is that the generalist model selects its strategies according to the reinforcement conditional on the conjunction of the two messages, just as in the traditional model. The difference is that the generalist model is not restricted in what the receiver can learn: the receiver is also learning about the atomic messages \emph{at the same time}. In this sense, the generalist model combines the advantages of both the minimalist and traditional model: the receiver is learning from all the information available. However, this comes at the expense of greater complexity.

\section{Objections}
\label{sec:challenges}

Before proceeding, we should stop to consider whether these moves, from the traditional signaling model to the minimalist or generalist models, is really legitimate. There are three immediate objections to consider. 

First, at face value, these models might seem unnatural, artificial or convoluted, in a way that traditional signaling game models are not.

For instance, the tempered softmax activation function used in the minimalist model might seem like an unnatural imposition. However, any signaling game architecture makes some choice of activation function, implicitly, or explicitly. For example, traditional signaling games trained with Roth–Erev learning also have an implicit linear activation function, in choosing the response in proportion to the reinforcement. We should not prefer linear activations a priori, and softmax offers one reasonable choice.\footnote{A wide variety of activation functions have seen use in machine learning contexts \citep{Goodfellow-neuralnets}. The appropriate choice of activation function is highly contingent, depending on the specifics we are trying to model. For example, softmax implements Boltzmann exploration, a standard model of stochastic choice in reinforcement learning and animal foraging; it therefore offers a cognitively and evolutionarily plausible way to bias the receiver toward its highest-valued action while still permitting exploration \citep{sutton2018reinforcement, Thrun1992}.} Plausibly, any monotone activation that accentuates the gap between the best- and second-best actions would be likely to yield similar qualitative outcomes.

However, I think the generalist model is \emph{more} natural than the traditional model. After all, there is no obvious reason for why a receiver should \emph{only} receive reinforcement based on whether their act corresponds to the full signal, rather than the atomic message components of the signal. In fact, there is something especially natural about a model which simply reinforces and learns about all the information available to the receiver without any restriction. Another way to put this is that evolutionary processes tend to explore the entire possibility landscape, so we should not restrict that landscape a priori. Furthermore, it is hardly surprising that a reinforcement learner will fail learn to interpret a signal compositionally if they can \emph{only} reinforce based on the whole signal, rather than the atomic messages that constitute its component parts. 

Second, perhaps there is a sense in which allowing the receiver to reinforce on the compositional parts of the signal might ``hard-wire'' the solution from the start. After all, we have created receivers who are potentially far more structured than the receivers in traditional signaling games. Perhaps the structure of the neural network in the minimalist model, or the implicit conditional probabilities in the generalist model already contain information about the solution?

However, these receivers exhibit more potential structure only because they are \emph{more general} types of learners. The potential structure exhibited by these receivers is not additional pre-learned structure or information. One way to see this is that the receivers of this type are not in any sense hard-wired to learning conjunctions of signals. These models could also be used to learn other logical operations such as inclusive or exclusive disjunctions.

Third, perhaps the receivers in these models are not really compositional at all. Perhaps, the final action taken by the receiver is still fully determined by the full conjunction of the messages, not by the atomic messages. In this sense, the lack of compositionality is just being hidden by an extra layer of complexity in the structure of the receiver.

This is easiest to see with the minimalist model. In effect the receiver has become a more sophisticated neural network agent, with two layers of beliefs, a first layer that contains the input message, and a second layer that selects an output action solely based on the conjunction of messages.  The final choice of action still depends solely on the conjunction of messages, just as in the traditional model. Likewise, in the generalist model, the receiver acts solely on the conjunction of messages, albeit whilst storing the other relevant information in a full joint probability distribution.

In one sense, this third objection is exactly right, but this is a bullet I am happy to bite. The final action does indeed depend on the full message, not on the component atomic messages. It is inevitable that such a receiver ultimately makes a decision based on the whole message: what is important is how the receiver goes about interpreting the atomic messages to come to their conclusion about which act to perform. The minimalist interprets the message compositionally but necessarily has intermediate steps of processing before reaching that final stage.  But in pushing the step \emph{inside} the receiver, we have allowed the receiver, \emph{taken as a whole}, to interpret messages compositionally. As such, the receiver as a whole can retain partial information and act accordingly when a sender's message is forgotten or replaced.

Likewise, the generalist receiver stores more sophisticated information, involving the individual probabilities and conditional probabilities. The relevant information about the action conditional on combinations \emph{is} stored, but so is store more complete information. Uniquely of all the models considered here, the generalist receiver effectively builds a model of the state of the world with \emph{all} of the information that is potentially available to them.

\section{Comparisons with Other Models}

As discussed in section \ref{sec:intro}, there are several alternative approaches to the solving the puzzle of compositionality, each carrying some similarities to the minimalist and generalist models proposed here. \citep{barrett2020compositional} propose three two-sender game models in which executive agents reinforce or suppress basic senders and enforce minimal signaling. In the special composition game, the two basic senders are pre-assigned one semantic dimension each. The role-free composition game removes those fixed roles. Reinforcement learning usually drives the senders to partition the state-space into complementary semantic dimensions. The general composition game keeps this role freedom but eliminates the explicit efficiency rule; instead every signal carries a cost that is deducted from pay-offs. Because unnecessary signals now reduce net reinforcement, the population again evolves the one-sender-per-dimension scheme across a wide range of cost/pay-off parameters. The executive sender (who decides which basic sender(s) may speak) and the executive receiver (who tells the basic receiver how to interpret what arrives) learn in tandem with the basic agents, so contextual demands plus either efficiency constraints or signal costs reliably steer the whole hierarchy toward a compositional communicative language. 

As such, receivers interpret each atomic message in isolation when the game’s context demands it; combining messages simply intersects the independently learned meanings. These models explain how compositionality might emerge as a result of linguistic and computational costs and how the sender might evolve to send only single terms when that is most efficient and hence be an active partner in the use of a compositional language. However, it is worth noting that these dynamics occur within individual learning agents, rather than through explicit population-level evolution. Note that the executive-functional architecture introduces significant complexity and relies on mechanisms that are not present in standard reinforcement-based agents. Moreover, these models arguably make the sender an active partner in optimizing for compositionality.

On the other hand, \citep{Franke2014} keeps the classic single-sender/single-receiver architecture but tweaks how reinforcement spreads. After a rewarded move, weight flows to similar states and similar symbols. Because each atomic signal’s reinforcement spills into new contexts, the receiver treats those atomic units as independently meaningful resources; when two familiar atomic units are sent together, their meanings combine predictably, so losing one still leaves the other’s contribution intact.  Together with a mild lateral-inhibition parameter, this spillover reinforcement learning can lead to compositional communication.

Finally, \citep{steinert2020emergence} shows that a variable-context signaling game,with multiple objects and gradable properties, coupled to neural-network learners and an attention mechanism produces stable compositional communication. One subset of signals encodes which dimension of a gradable property is relevant; another subset encodes the polarity (highest vs. lowest) along that dimension. The polarity signal modifies the dimension signal rather than contributing a second, independent proposition.

However, the minimalist and generalist models are in many respects architecturally simpler than the alternatives. After all, in many ways, both models are a natural generalization of the standard canonical Lewis–Skyrms signaling game, relaxing some of the assumptions about what information the receiver can learn. As such, the models require auxiliary learning mechanisms, no additional sender states, and no extra executive control elements. Likewise, the neural networks in the generalist receiver are substantially simpler than those required in the \citep{steinert2020emergence} (although more complex networks would surely be needed to interpret more sophisticated compositional signals).

As such, the minimalist and generalist models carry certain explanatory advantages, especially regarding the puzzle of compositionality. The puzzle arises because receivers in conventional signaling games only reinforce based on the entire message, rather than the components. The minimalist and generalist models confront this problem directly, by explicitly allowing reinforcement on the other signals, without introducing additional agents or learning mechanisms.

The minimalist and generalist models are nonetheless quite general. After all, the parameters governing the receiver can be varied or, in a more naturalistic model, tuned by evolutionary processes. More significantly, these models dispense with any artificial restriction on receivers to only learn on the full compositional messages: they are now rewarded for exploiting every statistical cue available to them. The generalist model is capable of approximating a full joint probability distribution over messages and acts, making it an obvious bridge to more Bayesian treatments of language evolution, whilst the minimalist receiver likewise suggests connections with classic work on neural-network learnability.

\section{Conclusions}

The receivers in traditional signaling game models do not interpret composite signals compositionally. However, we should never have expected them to do so. The receivers in such models explicitly learn and choose their action based on \emph{only} the entire message, not its components. They never consider information about the atomic messages. However, I have shown by example that we can construct signaling game models in which the receivers \emph{do} interpret signals compositionally. The key is that the receivers consider the component information available from the atomic messages.

Both of the compositional models are worthy of further consideration. The minimalist receiver \emph{only} considers the atomic messages, and requires a judicious choice of activation function in order to reach a signaling system. However, minimalist receivers of this type might provide a point of contact between signaling games and research involving neural networks. Indeed, these observations complement \citet{steinert2020emergence}, who uses larger recurrent neural networks models of agents to learn non-trivial compositional states. The arguments presented here suggest that these models are able to learn \emph{genuinely} compositional information, rather than simply appearing to do so.

As I discussed above, this is not surprising in retrospect. Artificial neural networks were developed precisely for the purpose of learning complex information, and their success in learning logical operators was demonstrated decades ago. The problems with learning compositional information in signaling games do not arise from the inherent structure of the game, but rather from the a priori restriction to very simple receivers, incapable of storing complex information with component parts. It seems likely that future progress in understanding the emergence of communication and language using signaling games will depend upon more sophisticated agents such as these.

The generalist receiver is a more naturalistic model. After all, why should we restrict the receiver to only take into account either the atomic  messages or the whole message? Simply allowing the receiver to consider \emph{all} of the information available to them brings some of the advantages of both approaches. Such a model is also more compatible with a Bayesian approach: after all, we can imagine such a receiver as gradually learning the full joint probability distribution via Bayesian learning methods rather than reinforcement learning. This seems likely to provide a fruitful towards developing more general or more explicitly Bayesian models of communication.

Either of the compositional approaches could serve as a starting point for more general investigations into signaling games and compositionality. In this paper, I have intentionally restricted attention to the simplest type of signaling game in which compositionality can be of relevance. However, there are obvious routes to generalize these models to cases where more complex information must be conveyed. The minimalist model could be elaborated by allowing more complex functions of the input information (for example allowing more general activation functions). The generalist model generalizes to more complex settings very naturally, by always allowing all possible beliefs to form and be reinforced.

Suppose that we needed to deal with very complex compositional data. Perhaps a more sophisticated minimalist receiver, suitable for the task, would be interpretable as a neural network with many hidden layers. Likewise, a more sophisticated generalist receiver would effectively learn an increasingly complex Bayesian network with more nodes. These speculative possibilities suggest some potentially promising avenues that deserve further inquiry.

\section*{Acknowledgements}
\noindent I am very grateful to Jeff Barrett for his generous and incisive assistance in clarifying the conceptual role and positioning of this paper. Discussions with Brian Skyrms, Cailin O'Connor, Brian Ball, Ioannis Votsis, Daniel Hermann, Tom Beevers, Alice Helliwell, Tom Williams, and Peter West have also inspired and contributed to the ideas developed here. Earlier versions of this work were presented at the Symposium on \emph{Games and Signalling} at the \emph{2024 Biennial Meeting of the Philosophy of Science Association} in New Orleans, where I benefited from helpful discussions with Brandon Boesch, Justin Bruner, Matthew Coates, Ryan Chen, Jack VanDrunen, and others. Finally, I would like to thank the two anonymous reviewers for their thoughtful comments, which substantially improved the clarity and focus of this paper.

\bibliographystyle{apacite}
\bibliography{disbib}
\end{document}